\documentclass[nohyperref]{article}

\usepackage{microtype}
\usepackage{graphicx}
\usepackage{svg}
\usepackage{ragged2e}
\usepackage{booktabs,tabularx}

\usepackage{subfigure}
\usepackage{booktabs} 

\usepackage{hyperref}

\usepackage[accepted]{icml2022}

\usepackage{amsmath}
\usepackage{amssymb}
\usepackage{mathtools}
\usepackage{amsthm}

\usepackage[capitalize,noabbrev]{cleveref}

\theoremstyle{plain}

\theoremstyle{definition}

\theoremstyle{remark}

\usepackage[textsize=tiny]{todonotes}

\icmltitlerunning{Transfer Learning with Joint Fine-Tuning for Multimodal Sentiment Analysis}

\begin{document}

\twocolumn[
\icmltitle{Transfer Learning with Joint Fine-Tuning for Multimodal Sentiment Analysis}



\begin{icmlauthorlist}
\icmlauthor{Guilherme Lourenço de Toledo}{usp}
\icmlauthor{Ricardo Marcacini}{usp}
\end{icmlauthorlist}

\icmlaffiliation{usp}{Institute of Mathematics and Computer Sciences (ICMC), University of São Paulo, São Carlos-SP, Brazil}

\icmlcorrespondingauthor{Guilherme Lourenço de Toledo}{guitld@usp.br}
\icmlcorrespondingauthor{Ricardo Marcacini}{ricardo.marcacini@icmc.usp.br}

\icmlkeywords{Machine Learning, Multimodal Learning, Deep Learning}

\vskip 0.3in
]



\printAffiliationsAndNotice{\icmlEqualContribution} 

\begin{abstract}
Most existing methods focus on sentiment analysis of textual data. However, recently there has been a massive use of images and videos on social platforms, motivating sentiment analysis from other modalities. Current studies show that exploring other modalities (e.g., images)  increases sentiment analysis performance. State-of-the-art multimodal models, such as CLIP and VisualBERT, are pre-trained on datasets with the text paired with images. Although the results obtained by these models are promising, pre-training and sentiment analysis fine-tuning tasks of these models are computationally expensive. This paper introduces a transfer learning approach using joint fine-tuning for sentiment analysis. Our proposal achieved competitive results using a more straightforward alternative fine-tuning strategy that leverages different pre-trained unimodal models and efficiently combines them in a multimodal space. Moreover, our proposal allows flexibility when incorporating any pre-trained model for texts and images during the joint fine-tuning stage, being especially interesting for sentiment classification in low-resource scenarios.
\end{abstract}

\section{Introduction}
\label{introduction}

Methods for sentiment analysis have been widely studied in recent years, both in academia and industry \cite{birjali2021comprehensive}. The key idea is to automatically identify sentiment polarities from data, such as texts and images, in order to analyze people's opinions and emotions about products, services, or other entities \cite{zhang2018deep}. Most existing methods focus on sentiment analysis on textual data \cite{poria2018multimodal}. However, recently there has been a massive use of images and videos on social platforms, motivating sentiment analysis from other modalities \cite{zhu2022multimodal}. Multimodal sentiment analysis was proposed to deal with these scenarios and combine the different modalities into more robust representations to improve sentiment classification.

A crucial step for multimodal sentiment analysis in real-world applications is to obtain sufficient training data, especially using state-of-the-art methods based on deep neural models. In unimodal scenarios, such methods already depend on large datasets for model training \cite{dang2020sentiment}. In the multimodal scenario, there is an extra challenge associated with the need to align instances of the different modalities \cite{zhu2022multimodal}. For example, a social media post must contain both the image and the associated text to form an instance in the multimodal scenario. Recent methods, such as CLIP \cite{radford2021learning} and VisualBERT \cite{li2019visualbert}, are pre-trained on datasets with the text paired with images. Although the results obtained by these models are promising, the need for multimodal pre-training is computationally expensive. On the other hand, dozens of pre-trained models for images (e.g., ResNet and Inception) and texts (BERT and Distilbert) have been proposed and made available for fine-tuning specific tasks, which require significantly less computational resources. Thus, we raise the following question: \textit{how to fine-tune different pre-trained unimodal models considering a multimodal objective for sentiment analysis tasks?}

This paper introduces a transfer learning approach using joint fine-tuning for sentiment analysis, where joint fine-tuning is a training technique considering multiple modalities. Pre-existing joint fine-tuning techniques assume that pre-trained models are originally multimodal \cite{yao2020morse}. On the other hand, our proposal is agnostic to initial unimodal pre-trained models. Moreover, we allow fine-tuning of both pre-trained models as a single loss function during the sentiment classifier training step. In practice, we are transferring knowledge from unimodal models that have been pre-trained from different (unpaired) image and text datasets. Our proposal is a deep neural network that incorporates two pre-trained models for both modalities (i.e., text and image). A fusion layer is added to the output of these models to project a latent space that unifies both modalities. This latent space is a multimodal feature extractor used to feed the output layer for sentiment classification. It is important to emphasize that the pairs of images and texts are only required in the fine-tuning step, and an attention mechanism is used to decide which modality should receive more weight during fine-tuning.

We carried out an experimental evaluation on two real datasets. The results show that our model is competitive with state-of-the-art models such as CLIP, even though it requires fewer resources to train the sentiment classifier. We also showed the importance of using both text and images for the sentiment analysis task by evaluating the respective unimodal sentiment analysis models. The source code of our project is publicly available\footnote{https://github.com/guitld/Transfer-Learning-with-Joint-Fine-Tuning-for-Multimodal-Sentiment-Analysis}.

\section{Related Works}

\subsection{Transfer Learning and Fine-Tuning}

Transfer learning aims to apply the knowledge learned in a general task to other related tasks \cite{zhuang2020comprehensive}. In deep learning, initially, a model is trained in a general context considering the existence of vast computational resources and then reused in another more specific task through fine-tuning, which uses significantly less computational resources.

The BERT model \cite{kenton2019bert} is famous for transfer learning and fine-tuning tasks in textual data. BERT fine-tuning has been widely explored in sentiment analysis for texts to identify aspects of an opinion (e.g., characteristic of a service or product) and classify sentiment polarity. Other models derived from BERT, such as DistilBERT, RoBERTa, and DeBERTa, have also been used. In particular, DistilBERT is a distilled and light version of BERT, maintaining similar performance but reducing approximately 40\% of the BERT parameters.

Regarding image sentiment analysis, deep neural models pre-trained on large image databases, such as ImageNet, are commonly employed in transfer learning and fine-tuning tasks --- usually from architectures based on convolutional neural networks. A pioneering study was presented  by \cite{campos2015diving}, which investigated different fine-tuning strategies for image sentiment analysis. The authors demonstrated that transfer learning and fine-tuning are promising strategies, surpassing state-of-the-art models. 


\subsection{Multimodal Learning}

While previous methods use only a single modality for training deep neural networks, recent work shows that sentiment analysis can be significantly improved by exploring the complementarities and correlations between visual and textual features \cite{zhu2022multimodal}.

Several methods have been proposed to jointly learn representations from images and texts. CLIP \cite{radford2021learning} is one of the main models for this multimodal learning, which explores pairs of texts and images in the pre-trained stage. During training, the goal is to predict the most relevant piece of text for a given image. Formally, CLIP tries to maximize the cosine similarity between pairs of images and texts in order to find the best alignment between visual and textual instances.

The representations learned by CLIP can be used for fine-tuning the most specific task. There are other multimodal networks for similar tasks, such as VisualBert \cite{li2019visualbert}, which is very popular for visual question answering.

\section{Joint Fine-Tuning for Sentiment Analysis using Pre-trained Unimodal Models}

The previously discussed multimodal models, such as CLIP and VisualBERT, require pairs of texts and images to learn the multimodal representations during the model pre-training. We propose a simple and effective alternative for multimodal sentiment analysis, which takes advantage of several pre-trained unimodal models and learns the multimodal representation only in the fine-tuning step.

\begin{figure}[!h]
\centering
\includegraphics[height=7cm]{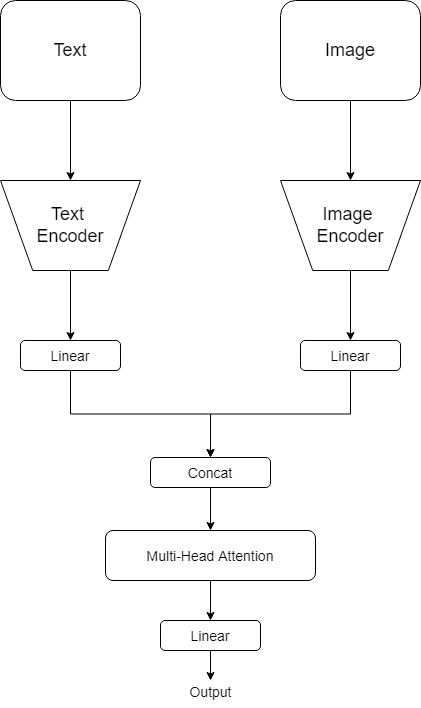}
\label{fig1}
\caption{Overview of the proposed architecture for joint fine-tuning with multiple modalities.}
\end{figure}

Figure 1 illustrates our proposed architecture for a pair of images and texts, here named $(x_t,x_v)$. Let $ENC_{T}(x_t)$ and $ENC_{V}(x_v)$ be two encoders resulting from pre-trained models, respectively for texts and images. Both $ENC_{T}(x_t)$ and $ENC_{V}(x_v)$ are unimodal models pre-trained on uncorrelated training data, for example BERT and ResNet, respectively.

Our architecture is agnostic to pre-trained unimodal models, which can have different dimensionality. For example, BERT-base encodes text into 768 features, while ResNet-18 encodes images into 512 features. Thus, the output of each encoder is processed by linear layers to project each modality in the same dimensionality size. A concatenate layer is responsible for unifying the feature vectors of each modality, followed by a Multi-Head Attention layer. This step combines the modalities and identifies features that are most relevant during the fine-tuning step via attention mechanisms.

The output layer uses the unified features to classify the sentiment of the $(x_t,x_v)$ pair. We use the categorical cross-entropy as the loss function for fine-tuning step. Note that although we have two different encoders $ENC_{T}(x_t)$ and $ENC_{V}(x_v)$, a single loss value is generated during fine-tuning, which is used to adjust the neural weights of both encoders via back-propagation.

We argue that there is a transfer learning from the pre-trained unimodal models during fine-tuning as part of the Forward Propagation step. After calculating the classification error, there is a joint fine-tuning of both encoders in the Backward Propagation step since we try to adjust the weights and biases using the same loss function and the same task.

\section{Experimental Evaluation}

\subsection{Datasets}

We evaluated our proposal on two multimodal datasets. The first dataset was provided by the MVSA (Sentiment Analysis on Multi-view Social Data) project \cite{niu2016sentiment}. The MVSA is composed of pairs of images and texts collected from Twitter and manually annotated in positive, neutral, and negative sentiment labels. For the experimental evaluation, we set up a sub dataset called MVSA-1.2k, which uses only pairs of images and texts where there was an agreement between most annotators regarding the labels. The MVSA-1.2K has 1200 samples from each label, totaling 3600 instances.

The second dataset is HatefulMemes \cite{kiela2020hateful}. This dataset contains 10k memes labeled as hateful or non-hateful, as presented by \cite{kiela2020hateful} who defines hateful as ``a direct or indirect attack on people based on characteristics, including ethnicity, race, nationality, religion, caste, sex, etc.''. Each meme has an image and associated text, and we use the labels (hateful or non-hateful) as a special case of sentiment analysis task.

\subsection{Experimental Setup}

To evaluate the performance of our proposal, we carried out a comparison with unimodal architectures, simple multimodal concatenation, and with the CLIP multimodal model, as described below:

\begin{itemize}
    \item ResNet: We used ResNet-50 as a unimodal sentiment classifier based only on images.

    \item Distilbert: a sentiment classifier based only on textual modality through a distilled version of BERT.
    
    \item SVM: concatenation of modalities embeddings generated by ResNet and Distilbert followed by a SVM model.

    \item ResNet + Distilbert: a simple concatenation of the two modalities followed by a feedforward network and softmax function for sentiment classification.

    \item CLIP: a multimodal model for texts and images, representing one of the state-of-the-art and competitive models for experimental comparison.

\end{itemize}

All models were fine-tuned in the classification task considering the two datasets presented above. A minimum of 3 epochs was used, followed by an early-stopping strategy based on the training loss scores. The AUC (Area under the ROC Curve) measure was used to compare the models.

\subsection{Results and Discussion}
Table 1 shows the experimental results considering 10-fold cross-validation.

\begin{table}[!htpb]
\scriptsize
        \caption{Experimental comparison of unimodal and multimodal models in relation to our proposal.}
        \label{tab:results} 
        \setlength{\tabcolsep}{4pt}
        \begin{tabularx}{8.5cm}{>{\hsize=1.1\hsize\bfseries\RaggedRight}X!{\extracolsep{\fill}}*{6}{>{\centering\arraybackslash\hsize=0.48\hsize}X}}
                \toprule[1pt]\midrule[0.3pt]
                \multicolumn{7}{c}{\textbf{Evaluation Results (AUC)}} \\ \midrule[0.3pt]
                & \multicolumn{2}{c}{\textit{MVSA-1.2k}} & \multicolumn{2}{c}{\textit{HatefulMemes}} \\
                \cmidrule(lr){2-3} \cmidrule(lr){4-5} \cmidrule(lr){6-7}%
                {Method} & Mean & Std & Mean & Std  \\
                \addlinespace%
                {SVM (concatenation)} & 0.6416 & $\pm{0.0001}$ & 0.5493 & $\pm{0.0002}$ \\
                \addlinespace%
                {ResNet} & 0.6383 & $\pm$0.0006 & 0.5359 & $\pm$0.0005  \\
                \addlinespace%
                {Distilbert} & 0.8011 & $\pm$0.0003 & 0.7298 & $\pm$0.0001  \\
                \addlinespace%
                {ResNet+Distilbert} & 0.7905 & $\pm$0.0025 & 0.6968 & $\pm$0.0004  \\
                \addlinespace
                {CLIP} & 0.8174 & $\pm$0.0004 & 0.7431 & $\pm$0.0005  \\
                \addlinespace
                {Our proposal} & 0.8131 & $\pm$0.0018 & 0.7461 & $\pm$0.0005  \\

                \midrule[0.3pt]\bottomrule[1pt]
        \end{tabularx}
\end{table}
\begin{table}[!htpb]
\scriptsize
        \caption{Aproximate Number of Parameters (in millions)}
        \label{tab:results2} 
        \setlength{\tabcolsep}{20pt}
        \centering
        \begin{tabularx}{8.5cm}{{}*6{>{\centering\arraybackslash\hsize=0.48\hsize}X}}
                \toprule[1pt]\midrule[0.3pt]
                \textbf{Method} & \textbf{Number of Parameters}  \\
                \addlinespace%
                \textbf{ResNet} & 23.5M  \\
                \addlinespace%
                \textbf{Distilbert} & 66.4M  \\
                \addlinespace%
                \textbf{ResNet+Distilbert} & 90.0M  \\
                \addlinespace
                \textbf{CLIP} & 152M \\
                \addlinespace
                \textbf{Our proposal} & 90.3M  \\

                \midrule[0.3pt]\bottomrule[1pt]
        \end{tabularx}
\end{table}

Note that CLIP is the state-of-the-art model with the best results, effectively combining multiple modalities to improve sentiment classification. However, the cost of pre-training this specific model requires extensive computational resources (see Table \ref{tab:results2}) and is significantly expensive for the sentiment analysis fine-tuning step due to its architecture. On the other hand, our proposal achieved competitive results using a much simpler fine-tuning strategy. Moreover, there is flexibility to incorporate any pre-trained unimodal models for text and image. For this experimental evaluation, we used the ResNet and Distilbert models in our proposal to facilitate the comparison with their unimodal versions for sentiment classification.

We also argue that simple concatenation strategies should be baselines for multimodal learning. Although it obtains slightly inferior results, its simplicity can be a good cost-effective alternative. In fact, our proposal can be seen as an extension of a concatenation strategy, but with the incorporation of an attention mechanism and joint fine-tuning. Finally, considering multimodal sentiment analysis, it is interesting to discuss the importance of each modality. Our experimental results with unimodal models show that texts can better discriminate the sentiment labels for the datasets used in this study. This result is automatically incorporated into our proposal via the attention mechanism that selects the most relevant features of each modality.

\section{Concluding Remarks}

Multimodal sentiment analysis is an increasingly relevant task due to human interaction growth in social networks. Its applications help extract opinions and sentiment from consumers and support the identification of hate speech and other types of violence in social networks. Deep learning methods have been used frequently for this task, mainly exploring textual features. More recent studies show that exploring other modalities, such as images, increases the performance of sentiment analysis.

State-of-the-art models for multimodal representations involving text and images are computationally expensive. We present an alternative and simpler architecture that maintains competitive performance compared to state-of-the-art models, such as CLIP. Directions for future work involve an extensive experimental evaluation of our approach considering different unimodal pre-trained models. We also plan to incorporate a multimodal explainability module as a strategy to identify the impact that each modality has on sentiment classification.

\bibliography{references.bib}

\begin{thebibliography}{13}
\providecommand{\natexlab}[1]{#1}
\providecommand{\url}[1]{\texttt{#1}}
\expandafter\ifx\csname urlstyle\endcsname\relax
  \providecommand{\doi}[1]{doi: #1}\else
  \providecommand{\doi}{doi: \begingroup \urlstyle{rm}\Url}\fi

\bibitem[Birjali et~al.(2021)Birjali, Kasri, and
  Beni-Hssane]{birjali2021comprehensive}
Birjali, M., Kasri, M., and Beni-Hssane, A.
\newblock A comprehensive survey on sentiment analysis: Approaches, challenges
  and trends.
\newblock \emph{Knowledge-Based Systems}, 226:\penalty0 107134, 2021.

\bibitem[Campos et~al.(2015)Campos, Salvador, Gir{\'o}-i Nieto, and
  Jou]{campos2015diving}
Campos, V., Salvador, A., Gir{\'o}-i Nieto, X., and Jou, B.
\newblock Diving deep into sentiment: Understanding fine-tuned cnns for visual
  sentiment prediction.
\newblock In \emph{Proceedings of the 1st International Workshop on Affect \&
  Sentiment in Multimedia}, pp.\  57--62, 2015.

\bibitem[Dang et~al.(2020)Dang, Moreno-Garc{\'\i}a, and De~la
  Prieta]{dang2020sentiment}
Dang, N.~C., Moreno-Garc{\'\i}a, M.~N., and De~la Prieta, F.
\newblock Sentiment analysis based on deep learning: A comparative study.
\newblock \emph{Electronics}, 9\penalty0 (3):\penalty0 483, 2020.

\bibitem[Kenton \& Toutanova(2019)Kenton and Toutanova]{kenton2019bert}
Kenton, J. D. M.-W.~C. and Toutanova, L.~K.
\newblock Bert: Pre-training of deep bidirectional transformers for language
  understanding.
\newblock In \emph{Proceedings of NAACL-HLT}, pp.\  4171--4186, 2019.

\bibitem[Kiela et~al.(2020)Kiela, Firooz, Mohan, Goswami, Singh, Ringshia, and
  Testuggine]{kiela2020hateful}
Kiela, D., Firooz, H., Mohan, A., Goswami, V., Singh, A., Ringshia, P., and
  Testuggine, D.
\newblock The hateful memes challenge: Detecting hate speech in multimodal
  memes.
\newblock \emph{NeurIPS}, 33:\penalty0 2611--2624, 2020.

\bibitem[Li et~al.(2019)Li, Yatskar, Yin, Hsieh, and Chang]{li2019visualbert}
Li, L.~H., Yatskar, M., Yin, D., Hsieh, C.-J., and Chang, K.-W.
\newblock Visualbert: A simple and performant baseline for vision and language.
\newblock \emph{arXiv:1908.03557}, 2019.

\bibitem[Niu et~al.(2016)Niu, Zhu, Pang, and Saddik]{niu2016sentiment}
Niu, T., Zhu, S., Pang, L., and Saddik, A.~E.
\newblock Sentiment analysis on multi-view social data.
\newblock In \emph{International Conference on Multimedia Modeling}, pp.\
  15--27, 2016.

\bibitem[Poria et~al.(2018)Poria, Hussain, and Cambria]{poria2018multimodal}
Poria, S., Hussain, A., and Cambria, E.
\newblock \emph{Multimodal sentiment analysis}.
\newblock Springer, 2018.

\bibitem[Radford et~al.(2021)Radford, Kim, Hallacy, Ramesh, Goh, Agarwal,
  Sastry, Askell, Mishkin, Clark, et~al.]{radford2021learning}
Radford, A., Kim, J.~W., Hallacy, C., Ramesh, A., Goh, G., Agarwal, S., Sastry,
  G., Askell, A., Mishkin, P., Clark, J., et~al.
\newblock Learning transferable visual models from natural language
  supervision.
\newblock In \emph{International Conference on Machine Learning}, pp.\
  8748--8763. PMLR, 2021.

\bibitem[Yao et~al.(2020)Yao, P{\'e}rez-Rosas, Abouelenien, and
  Burzo]{yao2020morse}
Yao, Y., P{\'e}rez-Rosas, V., Abouelenien, M., and Burzo, M.
\newblock Morse: Multimodal sentiment analysis for real-life settings.
\newblock In \emph{Proceedings of the 2020 International Conference on
  Multimodal Interaction}, pp.\  387--396, 2020.

\bibitem[Zhang et~al.(2018)Zhang, Wang, and Liu]{zhang2018deep}
Zhang, L., Wang, S., and Liu, B.
\newblock Deep learning for sentiment analysis: A survey.
\newblock \emph{Wiley Interdisciplinary Reviews: Data Mining and Knowledge
  Discovery}, 8\penalty0 (4):\penalty0 e1253, 2018.

\bibitem[Zhu et~al.(2022)Zhu, Li, Yang, Zhao, Liu, and Qian]{zhu2022multimodal}
Zhu, T., Li, L., Yang, J., Zhao, S., Liu, H., and Qian, J.
\newblock Multimodal sentiment analysis with image-text interaction network.
\newblock \emph{IEEE Transactions on Multimedia}, 2022.

\bibitem[Zhuang et~al.(2020)Zhuang, Qi, Duan, Xi, Zhu, Zhu, Xiong, and
  He]{zhuang2020comprehensive}
Zhuang, F., Qi, Z., Duan, K., Xi, D., Zhu, Y., Zhu, H., Xiong, H., and He, Q.
\newblock A comprehensive survey on transfer learning.
\newblock \emph{Proceedings of the IEEE}, 109\penalty0 (1):\penalty0 43--76,
  2020.

\end{thebibliography}
\bibliographystyle{icml2022}

\end{document}